\documentclass{article}
\usepackage{graphicx}
\usepackage{comment}
\usepackage{amsmath}
\begin{document}

\bibliographystyle{plain}
\pagenumbering{gobble}

\pagenumbering{arabic}
\pagestyle{empty}

\title{Attractor Image-Based Deep Learning of Arterial Pulse Waves for Age Classification}
\author{Sara Vardanega$^1$, Patrick Segers $^2$, Philip Aston $^{3,4}$, Ernst Rietzschel$^5$,\\ Jordi Alastruey $^{1*}$, Manasi Nandi $^{6*}$ \\ ** These authors contributed equally to this work \\}
\date{}
% leave an empty line between authors and affiliation

\maketitle
\thispagestyle{empty}
\vspace{0.5cm}
$^1$School of Biomedical Engineering and Imaging Sciences, King's College London, London, UK, \\
$^2$Institute of Biomedical Engineering and Technology, Ghent University, Ghent, Belgium, \\
$^3$National Physical Laboratory, Teddington, UK, \\
$^4$School of Mathematics and Physics, University of Surrey, Guildford, UK, \\
$^5$Department of Cardiovascular Diseases, Ghent University Hospital, Ghent, Belgium, \\
$^6$School of Cancer and Pharmaceutical Sciences, King's College London, London, UK\\\\
%\vspace{1cm}
Sara Vardanega, sara.vardanega@kcl.ac.uk \\
St Thomas' Hospital, Westminster Bridge Rd, London SE1 7EH\\

\begin{abstract}
Arterial pulse waveform morphology evolves with age, reflecting structural and functional changes in the cardiovascular system. Thus, vascular age is a valuable surrogate marker of cardiovascular health, and premature vascular ageing can indicate increased disease risk. Pulse wave analysis could support risk stratification in otherwise asymptomatic adults. We transformed pulse wave time-series data from photoplethysmography (PPG) and arterial tonometry into images, using the Symmetric Projection Attractor Reconstruction (SPAR) method. These SPAR images were used to train a convolutional neural network to classify healthy subjects into two closely spaced age groups (35–40 and 50–55 years). The model demonstrated consistent classification performance across internal and external test sets, achieving F1 scores above 70$\%$ for both PPG and tonometry signals. These results suggest that SPAR-derived pulse wave images contain discriminative morphological features even among healthy adults close in age. This proof-of-concept lays the groundwork for future research into the use of SPAR for early risk detection using smart wearables. 
\end{abstract}

\section{Introduction}
\par Vascular ageing (VA) is a complex process that involves the gradual deterioration of arterial structure and function over time, negatively impacting organ function \cite{climie2023vascular}. The gold-standard measurement for VA is carotid-femoral pulse wave velocity, but this requires trained personnel and is not routinely clinically available \cite{reshetnik2017oscillometric}. In healthy ageing, chronological and vascular ages typically correspond \cite{hamczyk2020biological}.

Deviations from this relationship, manifesting as premature VA, are associated with increased cardiovascular disease (CVD) risk. Therefore, early detection of premature VA is critical for timely prevention and management of CVD, which remains a leading global health burden.
Non-invasive pulse wave signals from photoplethysmography (PPG) or arterial tonometry can help assess vascular age, through analysis of pulse wave morphology, which changes with age. PPG is an optical method used in clinical and wearable devices to measure pulse waves at sites like the wrist and finger \cite{charlton2022}. Arterial tonometry, mainly used clinically, measures pressure from superficial arteries such as the radial or carotid arteries \cite{salvi2015noninvasive}. 
By comparing signal-based estimates of vascular age to a person's chronological age, we hypothesise we could identify early VA in community-based settings. 

\par This study focuses on image-based age classification leveraging PPG and tonometry-derived pulse waveforms, targeting two narrow yet clinically relevant age cohorts (35–40 and 50–55 years), representing a critical window during which individuals may begin to manifest subclinical CVD risk.
Pulse waves were converted into images using the Symmetric Projection Attractor Reconstruction (SPAR) method, which condenses time-series data into a single image \cite{aston2018beyond}. These images were used to train a convolutional neural network (CNN) to classify age, assigning each unseen image to the most likely of the two classes ($\leq $40 or $\geq$50).

 \begin{figure*}[h!]
    \centering
    \includegraphics[width=0.8\textwidth]{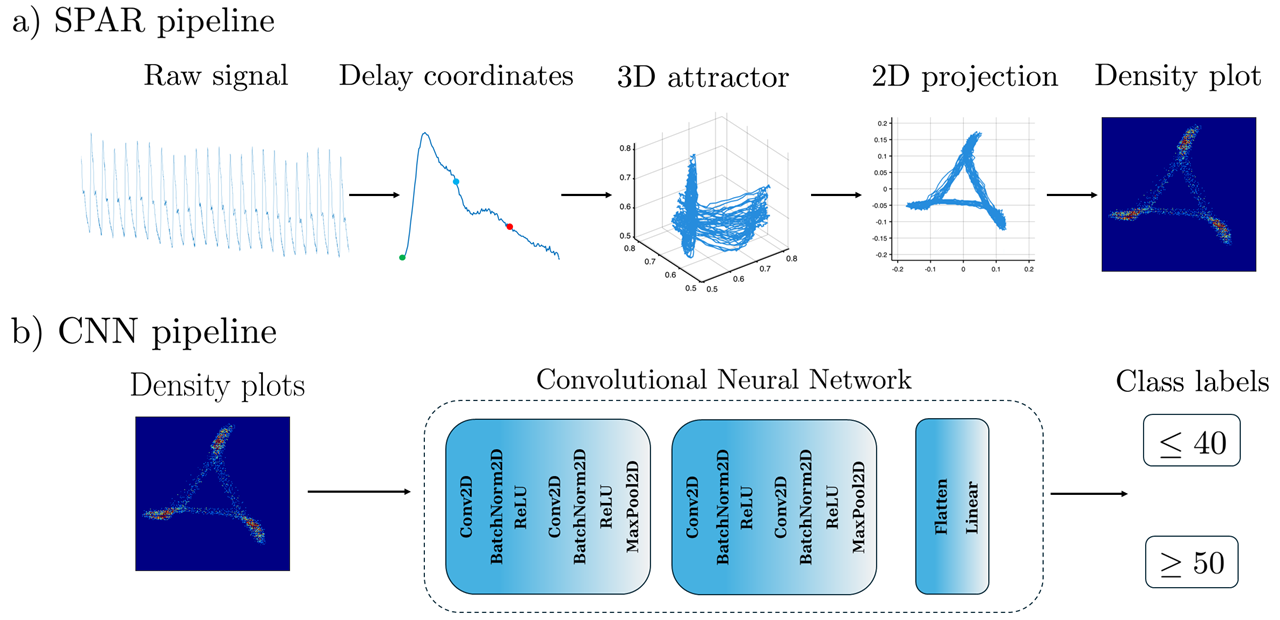}
    \caption{Full pipeline adopted in this study. Section a) illustrates the SPAR pipeline, which processes a raw signal into a SPAR attractor image (density plot). Section b) illustrates the CNN pipeline, where CNN takes as input the density plots and gives as output a class label.}
    \label{pipeline figure}

\end{figure*}
\section{Methods}

\subsection{Datasets}
 Two datasets were used in this study: Round 1 data from the Asklepios Study and the Vortal dataset (Table \ref{asklepios vortal population composition table}).\\
Both comprised recordings from participants free from diagnosed CVD at study initiation, collected in the supine position. 
The Asklepios Study \cite{rietzschel2007rationale} includes 2,524 individuals (30–59 years, 52$\%$ female) randomly sampled from two twinned Belgian communities. The exact chronological age of each participant was labelled. Arterial tonometry waveforms (20 s, 200 Hz) were acquired at a single centre by one trained operator using the same device. The Vortal dataset \cite{charlton2016assessment} contains finger PPG recordings from 56 subjects: 40 labelled as 'Young' (18–35 years, 53$\%$ female) and 16 as 'Elderly' (70+ years, 56$\%$ female). Data were collected in a London clinical trials unit (approx. 10 min per subject, 125 Hz). 
%\vspace{-0.4cm}
\subsection{Data selection}
 From the original Asklepios population, a subgroup was considered for this study.
We included only participants aged 30-40 and 50-59 years. 
Obese subjects (BMI $\geq$ 30 kg/m$^2$) and those with high blood pressure (systolic $\geq$ 140 mmHg or diastolic $\geq$ 90 mmHg) were excluded, following the 2024 ESC guidelines for hypertension classification \cite{mcevoy20242024}. For the Vortal dataset, we used the whole cohort, as no metadata were available.
The characteristics of both populations, assumed to be healthy, are shown in Table \ref{asklepios vortal population composition table}.
\begin{table}[htbp]
\caption{ Population characteristics for each dataset used in this study.}\label{asklepios vortal population composition table}
\vspace{4 mm}
\centerline{\begin{tabular}{lcr} \hline\hline
& \textbf{Asklepios} & \textbf{Vortal}\\ \hline
 No. of participants & 2,524& 56    \\

Sex (M/F) & 1,223/1,301& 27/29  \\
 Age range (years) & 30-59 & 18-35,70+  \\
 Signal type & Tonometry & PPG   \\         

Raw signal length & 20 s & 10 min \\
Analysed signal length & 20 s & 20 s \\
 \hline\hline
\end{tabular}}

\end{table}
\begin{table}%[htbp]
\caption{Total and selected number of subjects in the Asklepios and Vortal datasets, along with the number of SPAR images used in each group.}\label{Asklepios and vortal population table}
\vspace{4 mm}
\centerline{\begin{tabular}{lccr} \hline\hline
\textbf{Cohort} & \textbf{Total}& \textbf{Selected }   & \textbf{Images}\\ \hline

 Asklepios 30-34 & 15 & 12  & 12\\
 Asklepios 35-40 & 456 & 341  & 341\\
 Asklepios 50-55 & 514 & 251   & 251\\
 Asklepios 56-59 & 160 & 77  & 77 \\
 Vortal 18-35 & 40 & 40 & 1302 \\
 Vortal 70+ & 16 & 16 & 528 \\
 \hline\hline
\end{tabular}}

\end{table}

\begin{figure*}
    \centering
    \includegraphics[width=0.8\linewidth]{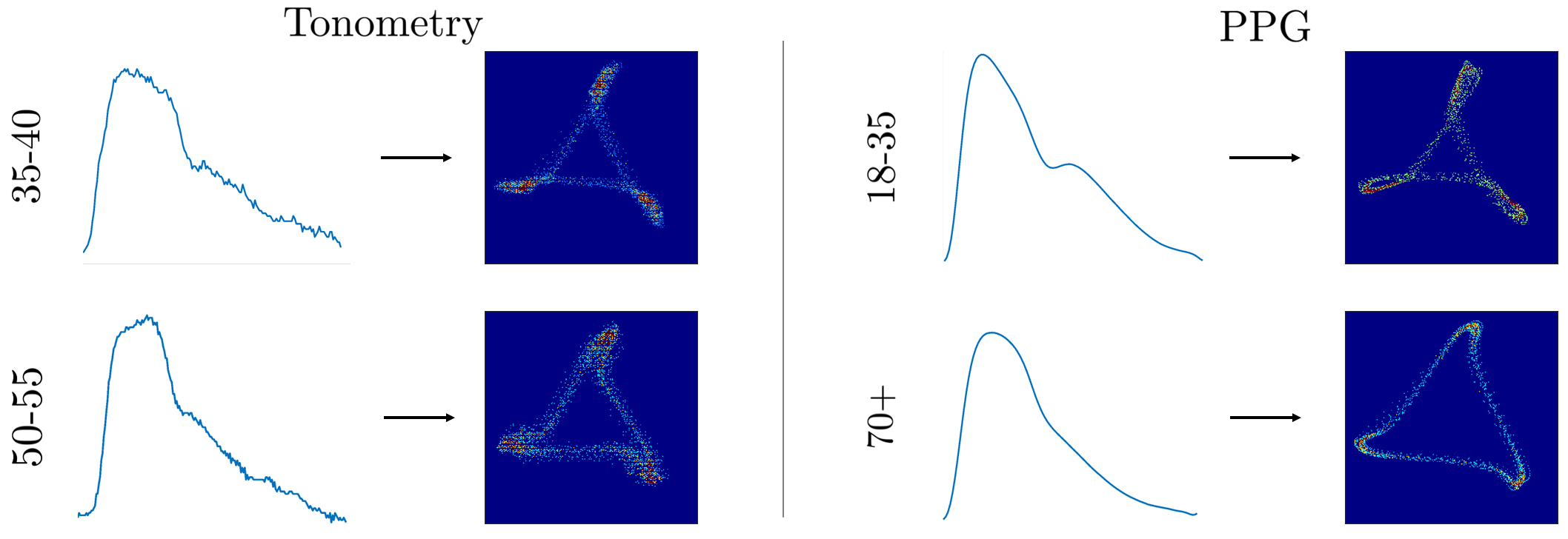}
    \caption{Examples of single tonometry (left) and PPG (right) pulse waveforms and their correspondent SPAR attractor images, generated using 20 seconds of data.}
    \label{results figure ppg vs tonometry}
\end{figure*}

\subsection{Symmetric Projection Attractor Reconstruction (SPAR) method}
The SPAR method \cite{aston2018beyond} is a non-fiducial points-based method that combines mathematics and cardiovascular physiology to quantify pulse waveform morphology and variability.  Given a raw pulse wave signal, the average cycle length is first estimated for the selected time window. Next, a three-dimensional (3D) attractor is generated using three delay coordinates, each separated by one-third of the average cycle length. The 3D attractor is then projected onto a two-dimensional (2D) plane normal to the unit vector and converted into a density plot (Figure \ref{pipeline figure}a). This plot is the final attractor image used to train and validate CNNs in this work.\\
For the Asklepios dataset, we used one 20-second segment per subject, resulting in one attractor image per participant. For the Vortal dataset, we extracted multiple non-overlapping 20-second segments from the 10-minute recordings, generating around 30 images per subject. 
\subsection{Model and metrics}
The CNN used in this study was based on a simplified TinyVGG architecture originally presented by Wang et al. \cite{wang2020cnn}. It comprises two convolutional blocks followed by a final classification layer (Figure \ref{pipeline figure}b). \\ Model performance was evaluated using sensitivity (TP / (TP + FN)), specificity (TN / (TN+FP)) and the F1 score, which is the harmonic mean of precision (TP / (TP + FP)) and sensitivity; with TP being the true positives, TN the true negatives, FP the false positives and FN the false negatives.
In this study, the positive class corresponded to the 50-55 years group, and the negative class to the 35-40 years group. Therefore, sensitivity measured the model's ability to correctly identify subjects aged $\geq$ 50 years, while specificity measured its ability to correctly classify those aged $\leq$ 40 years.
\subsection{Training and testing }\label{train, validation and test subsection}
The model was trained on 80\% of the Asklepios data from the 35-40 and 50-55 age groups. The remaining 20\% was split equally into a validation set (10\%) and  a test set (10\%). Given the relatively small size, 10-fold cross-validation was performed to improve robustness, thus obtaining 10 different sets of model weights which were evaluated separately when testing. The model was then tested on different test sets, reported in Table \ref{Asklepios and vortal population table}. 
When testing on the larger Asklepios population, the same 10-fold structure was maintained to minimise bias. The Vortal test set remained constant across all evaluations. 
\section{Results}\label{Results section}
The model reached average F1 scores of at least 70\%, with sensitivity $>$67\% and  specificity $>$79\% across all test sets (Table \ref{Results table}). Overall, performance improved with larger test sets and broader age ranges, suggesting robust model generalisation. The higher sensitivity observed in these sets indicates greater accuracy in classifying older subjects. Additionally, standard deviations decreased with broader test sets, reflecting more consistent performance across cross-validation folds. \par When evaluated on PPG signals, the model showed higher F1 scores compared to the baseline test on the Asklepios dataset, together with an increase in sensitivity and a slight decrease in specificity. These results highlight the ability of the model, combined with the SPAR method, to detect age-related morphological changes across both tonometry and PPG signals. Figure \ref{results figure ppg vs tonometry} illustrates these morphological differences. In younger individuals (35-40 years for tonometry, 18-35 years for PPG), pulse waveforms display a distinct secondary peak, resulting in attractors with “looped” edges and closed centres. In contrast, older subjects (50-55 years for tonometry, 70+ years for PPG) exhibit attenuated or absent secondary peaks, producing more open attractors with reduced looping, highlighting age-related waveform changes.

\begin{table}
\caption{Model performance across different test sets.}\label{Results table}
\vspace{4 mm}
\centerline{\begin{tabular}{lccr} \hline\hline
& \textbf{F1 score (\%)} & \textbf{Sens (\%)} & \textbf{Spec (\%)} \\ \hline
Test Asklepios   & 70.9 $\pm$ 8.6  & 67.0 $\pm$ 12.3 & 85.0 $\pm$ 6.3  \\
35-40 and 50-55 & & & \\
Test Asklepios   & 79.3 $\pm$ 2.0 & 70.5 $\pm$ 3.1 &  84.3 $\pm$ 4.8  \\
30-40 and 50-59 & & & \\
Test Vortal & 72.8 $\pm$ 2.5& 86.9 $\pm$ 5.9 & 79.0 $\pm$ 2.0 \\
 \hline\hline
\end{tabular}}

\end{table}
\section{Discussion and conclusion}
We have presented a simple yet effective model that can classify chronological age within a healthy population free from CVD, using two non-invasive pulse wave signals: arterial tonometry and PPG.  We purposefully focused on healthy individuals for whom chronological and vascular ages are expected to align. The strong classification performance achieved on both tonometry and PPG test sets shows that each signal type contains enough morphological information to distinguish between the two selected age groups. These results suggest that the model successfully captures age-related changes in the cardiovascular system that manifest as alterations in pulse wave morphology. \par Importantly, the presence of noise in the tonometry recordings (Figure \ref{results figure ppg vs tonometry}) did not impact the quality of the resulting attractor images, when compared to those derived from PPG. This indicates that the SPAR method is inherently robust to signal noise, preserving relevant morphological features despite variability in signal acquisition quality.\par The main limitation of this work lies in the small size of the training set, which was mitigated through 10-fold cross-validation to improve model robustness and generalisability. Furthermore, the detailed selection criteria applied to the Asklepios dataset could not be applied to the Vortal dataset due to the absence of detailed metadata. As a result, the Vortal dataset may include individuals with high blood pressure or BMI. However, the large age difference between the younger and older Vortal groups supports its suitability as a proof-of-concept PPG test set for age classification.
\par Model performance could be increased by implementing a more complex CNN model and by training on a larger and more diverse population. This would allow for evaluation of the interplay between model complexity and performance. Whilst CNNs can be directly applied to raw pulse waves signals, SPAR’s robustness on noisy signals and at-a-glance summary of multiple pulse waves as a single 2D image may be more intuitive for end users for visual interpretation.
\par In summary, while age-related differences in pulse wave morphology are well established, this study demonstrates a SPAR-based CNN approach capable of classifying individuals into two closely spaced age ranges within a CVD-free population. Moreover, our findings indicate that tonometry and PPG signals are sufficiently similar to enable this classification task. Given the infrastructure demands of gold-standard VA assessments such as pulse wave velocity, our findings support further research into the use of SPAR-transformed PPG signals from community or wearable devices for the early detection and stratification of cardiovascular risk.

\section*{Acknowledgments}  

The project (22HLT01 QUMPHY) has received funding from the European Partnership on Metrology, co-financed from the European Union’s Horizon Europe Research and Innovation Programme and by the Participating States. Funding for King’s College London was provided by Innovate UK under the Horizon Europe Guarantee Extension, grant number 10087011.
\bibliography{main}

\end{document}